\documentclass[11pt, a4paper, logo, copyright, nonumbering]{googledeepmind}

\pdftrailerid{redacted}

\makeatletter
\renewcommand\bibentry[1]{\nocite{#1}{\frenchspacing\@nameuse{BR@r@#1\@extra@b@citeb}}}
\makeatother

\usepackage{kantlipsum, lipsum}
\usepackage{dsfont}
\usepackage{gdm-colors}
\usepackage{float}
\usepackage{multicol}
\usepackage{enumitem}

\usepackage[authoryear, sort&compress, round]{natbib}

\graphicspath{{figures/}}

\newcommand{\name}{\textit{ALOHA 2} }
\newcommand{\namecomma}{\textit{ALOHA 2}}
\newcommand{\aloha}{\textit{ALOHA} }
\newcommand{\alohacomma}{\textit{ALOHA}}

\newcommand{\figureref}[1]{\hyperref[#1]{Figure~\ref*{#1}}}

\definecolor{codegreen}{rgb}{0,0.8,0}

\title{ALOHA 2: An Enhanced Low-Cost Hardware for Bimanual Teleoperation}

\correspondingauthor{aloha-2@googlegroups.com}

\reportnumber{} % Leave blank if n/a

\vspace{-0.5cm}
\author[ ]{ALOHA 2 Team\hspace{-0.5ex}}
\author[1]{Jorge Aldaco}
\author[1]{Travis Armstrong}
\author[1]{Robert Baruch}
\author[1]{Jeff Bingham}
\author[1]{Sanky Chan}
\author[1]{Kenneth Draper}
\author[1]{Debidatta Dwibedi}
\author[1,2]{Chelsea Finn}
\author[1]{Pete Florence}
\author[1]{Spencer Goodrich}
\author[1]{Wayne Gramlich}
\author[1]{Torr Hage}
\author[1]{Alexander Herzog}
\author[1]{Jonathan Hoech}
\author[1]{Thinh Nguyen}
\author[1]{Ian Storz}
\author[1]{Baruch Tabanpour}
\author[1,3]{Leila Takayama}
\author[1]{Jonathan Tompson}
\author[1]{Ayzaan Wahid}
\author[1]{Ted Wahrburg}
\author[1]{Sichun Xu}
\author[1]{Sergey Yaroshenko}
\author[1]{Kevin Zakka}
\author[1,2]{Tony Z. Zhao}

\affil[]{Google DeepMind}
\affil[2]{Stanford University}
\affil[3]{Hoku Labs}
\affil[ \hspace{-0.6ex}]{Authors listed in alphabetical order, with contributions listed in Appendix.}
\begin{abstract}
Diverse demonstration datasets have powered significant advances in robot learning, but the dexterity and scale of such data can be limited by the hardware cost, the hardware robustness, and the ease of teleoperation. We introduce \namecomma, an enhanced version of \aloha that has greater performance, ergonomics, and robustness compared to the original design.
To accelerate research in large-scale bimanual manipulation, we open source all hardware designs of \name with a detailed tutorial, together with a MuJoCo model of \name with system identification. See the project website at \url{aloha-2.github.io}.
\end{abstract}

\begin{document}

\maketitle

\begin{figure}[H]
	\centering
	\includegraphics[width=\columnwidth,height=0.55\textheight,keepaspectratio]{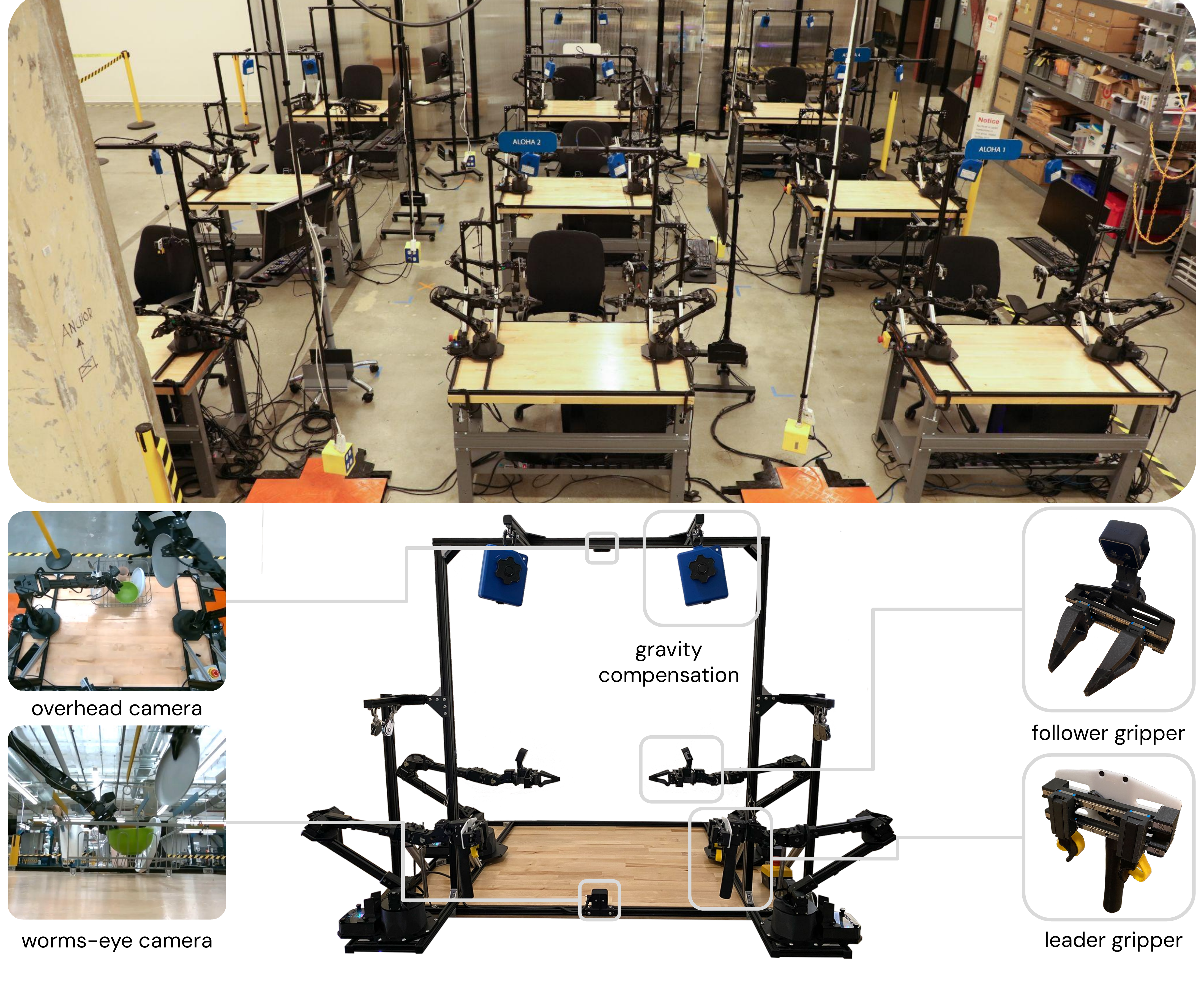}
	\vspace{-0.5cm}
	\caption{\textit{\textbf{\namecomma: An Enhanced Low-Cost Hardware for Bimanual Teleoperation.}} \textit{\textbf{Top:}} The fleet of \name robots capable of collecting 1000s of demonstrations per day. \textit{\textbf{Bottom:}} A detailed image of an \name workcell with gravity compensation, the redesigned leader and follower grippers, and images from the frame-mounted cameras.}
	\label{fig:figure1}
\end{figure}

\section{ALOHA 2}

\namecomma, like the original \cite{zhao2023learning}, consists of a bimanual parallel-jaw gripper workcell with two ViperX 6-DoF arms \citep{ViperX} (the "follower"), along with 2 smaller WidowX arms  \citep{WidowX} (the "leader"). The WidowX contains the same kinematic structure as the ViperX, in a smaller form factor. The follower joints are synchronized with the leader arms, and users teleoperate the follower arms by backdriving, or "puppeteering", the leader arms. The setup also contains cameras that produce images from multiple viewpoints, allowing collection of RGB data during teleoperation. The robots are mounted on a 48" x 30" table, with an aluminum cage to provide additional mount points for cameras and a gravity compensation system, which are detailed below. 

To support research on complex manipulation tasks, we aim to significantly scale up data collection on the \aloha platform, including the number of robots in use, the amount of hours of data per robot, and the diversity of data collection.  %and ensure long term usefulness of the data.
This scaling process shifts the requirements and scope relative to the first \aloha platform.
For \namecomma, we build on the strengths of \aloha platform while also targeting the following areas for further improvement:

\begin{itemize}
    \item \textbf{Performance and Task Range}: We seek to enhance key components that enable \alohacomma's performance, including grippers and controllers, to enable a broader set of manipulation tasks.% and provide performance closer to that of more advanced, but costlier, robotic platforms. pete: i commented this out because i don't know of any better costlier robot platforms!
    \item \textbf{User Friendliness and Ergonomics}: To optimize data collection at scale, we prioritize user experience and comfort. This includes improvements to the responsiveness and ergonomic design of user-facing systems.
    \item \textbf{Robustness}: We want to increase system robustness to minimize downtime caused by diagnosis and repairs.  This involves simplifying mechanical designs and ensuring the overall ease of maintenance for a larger robot fleet.
\end{itemize}

To this end, we make the following concrete improvements:

\begin{itemize}
	\item \textbf{Grippers}: We create a new low-friction rail design for both the leader and follower grippers. For the leader robots, this improves teleoperation ergonomics and responsiveness. For the followers, this improves latency and the force output of the grippers.
	In addition, we upgrade the grip tape material on the fingers to improve durability and grasping of small objects. 
% 	These changes enable higher quality teleoperation and a higher diversity of tasks.
	\item \textbf{Gravity Compensation}: We create a passive gravity compensation mechanism using off-the-shelf components that  improves the durability compared to \aloha's original rubber band system.
% 	to reduce user wear during teleoperation, 
% 	allowing collection of higher quality long horizon demonstrations.
	\item \textbf{Frame}: We simplify the frame surrounding the workcell, while maintaining the rigidity of the camera mounting points. These changes allow space for both human-robot collaborators and props for the robot to interact with.
	\item \textbf{Cameras}: We use smaller Intel RealSense \citep{keselman2017intel} D405 cameras and custom 3D-printed camera mounts to reduce the footprint of the follower arms, which less inhibits manipulation tasks.  These cameras also have a larger field of view, provide depth, have global shutter, and allow for more customization compared to the original consumer-grade webcams.
	\item \textbf{Simulation}: We model the exact specifications of the \name robot in a MuJoCo model in MuJoCo Menagerie, which allows improved data collection, policy learning, and evaluation in simulation for challenging manipulation tasks.
\end{itemize}

We find that these improvements make it easier to teleoperate challenging tasks tasks like folding a T-shirt, tying a knot, throwing objects,
or industrial tasks with tight tolerances. These improvements make it easier to collect 100s of demonstrations on these tasks per robot per day.

\section{Hardware}

\begin{figure}[t]
	\centering
	\includegraphics[width=0.9\columnwidth]{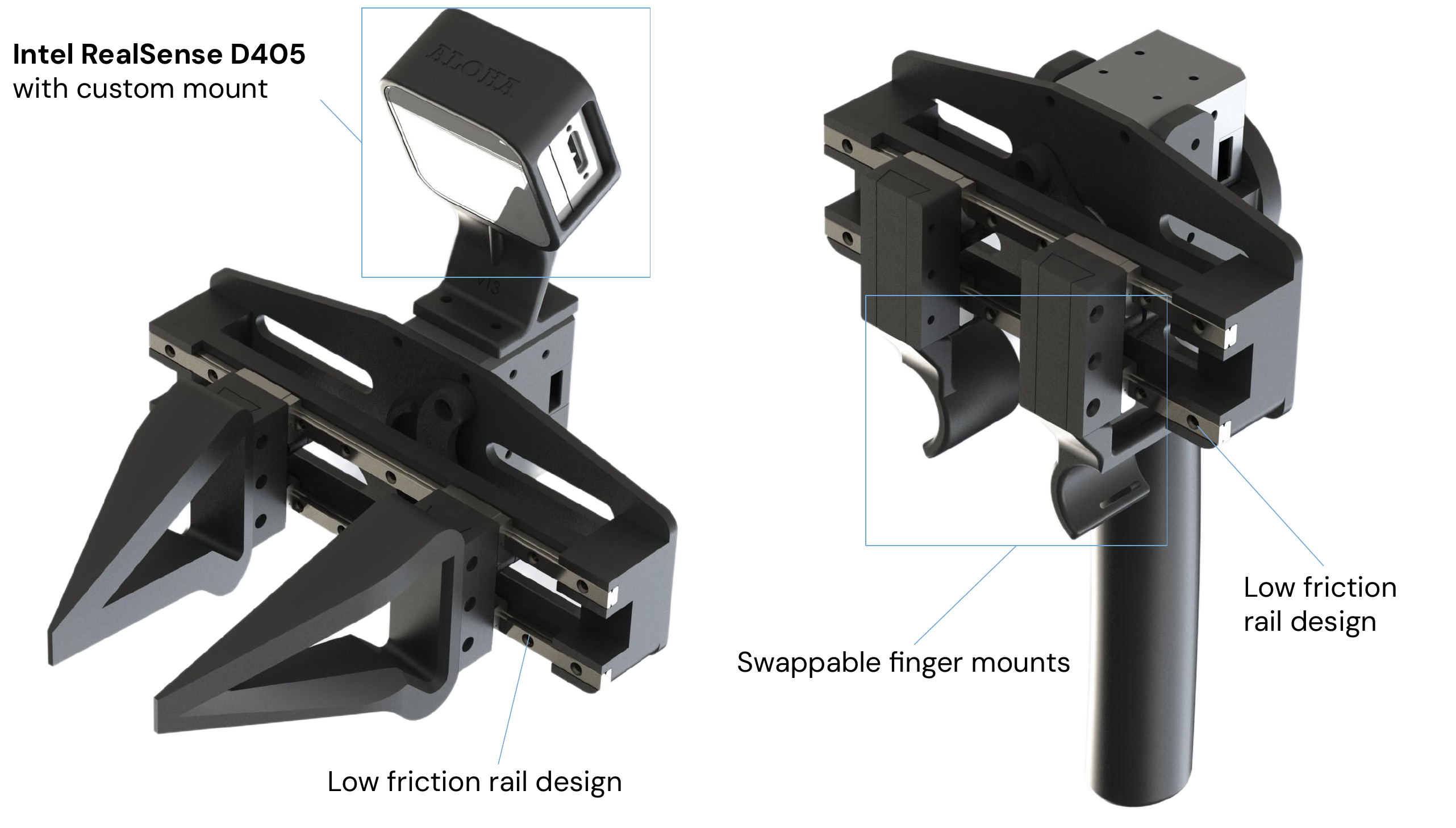}
	\caption{\textit{\textbf{Renderings of leader and follower gripper designs.}} \textit{\textbf{Left:}} follower gripper with Intel RealSense D405, custom 3d-printed camera mount, and low friction follower rail design. \textit{\textbf{Right:}} leader gripper with swappable finger mounts and low friction leader rail design.}
	\label{fig:gripper_leader_follower}
\end{figure}

\subsection{Leader Grippers \normalfont{\textit{(The device the human holds in their hands)}}}

\begin{figure}[t]
	\centering
	\vspace{-0.3cm}
	\includegraphics[width=.4\columnwidth]{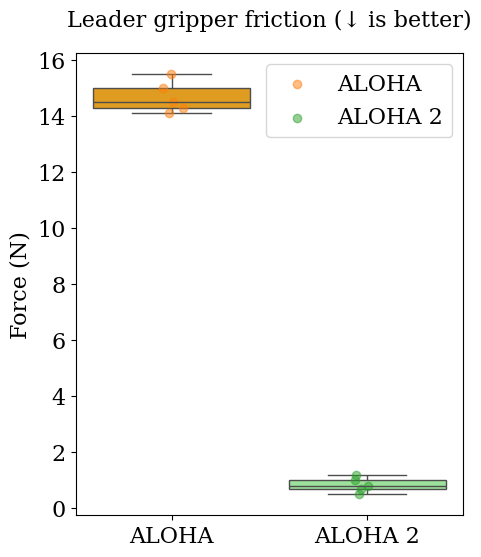}
	\includegraphics[width=.4\columnwidth]{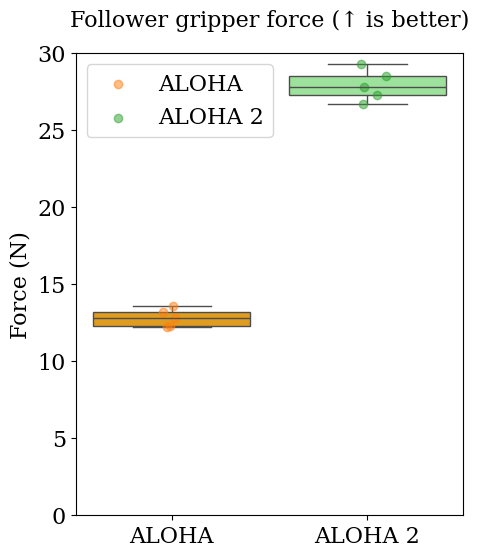}
	\caption{\textit{\textbf{\name improves the ergonomics and closing force of grippers. }}\textit{\textbf{Left:}} Force required from the operator to open leader grippers. \name reduces the force from 14.68N to 0.84N, reducing hand fatigue and improving ergonomics. \textit{\textbf{Right:}} Closing force at the tip of follower grippers. \name is capable of exerting more than double the force compare to the old design (27.9N vs. 12.8N)}.
	\label{fig:resistance}
\end{figure}

For smoother teleoperation and improved ergonomics, we replace the original scissor leader gripper design from \aloha with a low friction rail design with reduced mechanical complexity.  To further reduce strain, we also lower the backdriving friction by swapping the original leader's gripper motors (XL430-W250-T) with a lower-friction alternative (XC430-W150-T), which have a lower gear ratio and use lower-friction metal instead of plastic gears.\footnote{ \url{https://www.youtube.com/watch?v=JwAkSwwa0A4}} The new design requires approximately 10 times less force to open and close than the previous \aloha scissor design (Fig.~\ref{fig:resistance}). The lower friction significantly reduces the operator's hand fatigue and strain during long data collection sessions, notably on the lumbrical muscles \footnote{\url{https://my.clevelandclinic.org/health/body/25060-anatomy-of-the-hand-and-wrist}} responsible for opening the gripper.

\begin{figure}[t]
	\centering
	\includegraphics[width=\columnwidth]{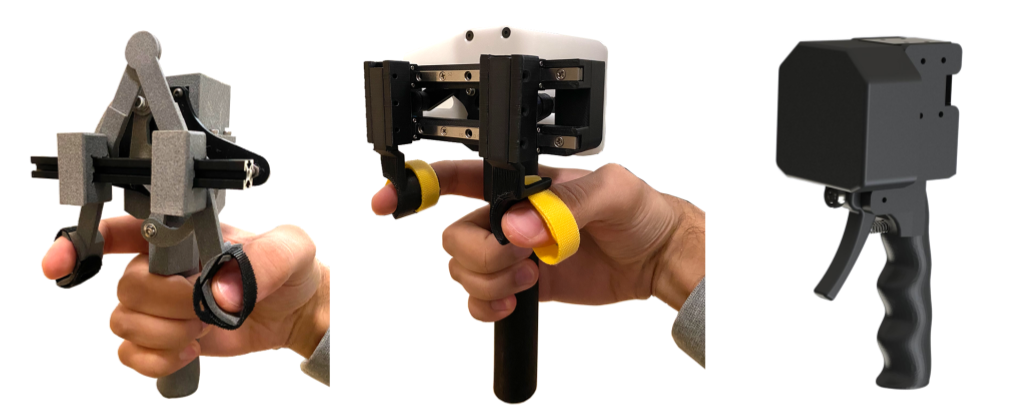}
	\caption{\textit{\textbf{The leader grippers evaluated for \namecomma.}} \textit{\textbf{Left:}} the original \aloha scissor design. \textit{\textbf{Middle:}} low friction rail design which was ultimately chosen based on user studies. \textit{\textbf{Right:}} rendering of the trigger design that was also considered.}
	\label{fig:leader_grippers}
\end{figure}

When deciding on the linear rail design, we also considered two additional designs. First is the original \aloha scissor design, which uses custom 3d printed rotors and rails to backdrive the leader gripper motor. In addition, we evaluate a spring loaded trigger design, where pulling the trigger closes the gripper and releasing it opens the gripper to it's neutral open position (See \figureref{fig:leader_grippers} for images and renderings of the evaluated grippers). We had 6 users teleoperate \name to unwrap candy using the original \aloha scissor design, the linear rail design, and the trigger design. While users had varying preferences, the linear rail was rated well by nearly all operators.

\subsection{Follower Grippers \normalfont{\textit{(The end-effector of the robot, i.e. the robot fingers that manipulate objects)}}}

We design and manufacture low-friction follower grippers, replacing the original design from \alohacomma. The lower friction reduces perceived latency between leader and follower grippers, significantly improving user experience during teleoperation. We show the difference in latency between the old design and new design in the supplementary video. The new grippers are also capable of applying 2 times more force than the old design, allowing for stronger and more stable grasps of objects.

In addition, we improve the compliance of the gripper mechanism by removing the original PLA + acrylic structure, and replacing it with 3D printed carbon fiber nylon. Both the gripper fingers and the supporting structure can deform when loaded, improving the safety of the system.

We keep the "see-through" design of the finger links from the original \alohacomma. In addition, we make improvements to the gripping tape on the fingers. We find that the original gripping tape on \aloha wears out over time, and the roundness at the tips of the fingers makes picking up small objects difficult. We apply a polyurethane gripping tape to the inside of the gripper. 
%%CF.1.24: more info on what this "new type of tape" is?
We also apply strips of tape on the outside of the finger to increase traction for tasks that require manipulating objects with the outer side of the gripper.

\subsection{Gravity Compensation}

We design a more robust \textit{passive} gravity compensation for the leader robots to ease operator wear during teleoperation. We construct this using off-the-shelf components, including adjustable hanging retractors that allows operators to tune the load balancing forces to their comfort level. 

We run a study to evaluate the \textit{passive} hardware gravity compensation system against an \textit{active} software-based system. We develop the active system using inverse dynamics from the MuJoCo model to calculate equivalent torques for the gravity load and then command these torques to the joints of the leader robot. To conduct the study, 6 users teleoperate the robots for 10 minute sessions, and attempt to perform a precise task of inserting shapes into corresponding holes in a box. The operators attempt the task on both systems in a randomly assigned order.
We compare performance on the task across the two systems, and find that on average operators performed better with the passive gravity compensation system (1.38 vs 0.97 shapes inserted per minute). Based on feedback from study participants, we speculate that the passive system allows more smooth and predictable movements. 
Study participants mention that the active system requires more force and results in choppier movement, perhaps due to poorly tuned servo gains or slight latency in applying counteracting forces. In addition, we find that the passive system provides two additional advantages. First, it can allow safer teleoperation by fully disabling joint actuation on the leaders,
as software bugs or edge cases may cause large or unexpected movements of the leader robots. Second, the force retractors give a natural centering for the arms to keep the wrists on the leader arms from over-rotating, which seemed to be a weakness of the active gravity compensation. Despite choosing the passive system for \namecomma, we speculate that an active system can perhaps be developed and tuned to surpass performance of our passive system, and could potentially be extended to allow useful features like providing tactile feedback to the user.

\subsection{Frame}

\begin{figure}[t]
	\centering
	\includegraphics[width=\columnwidth]{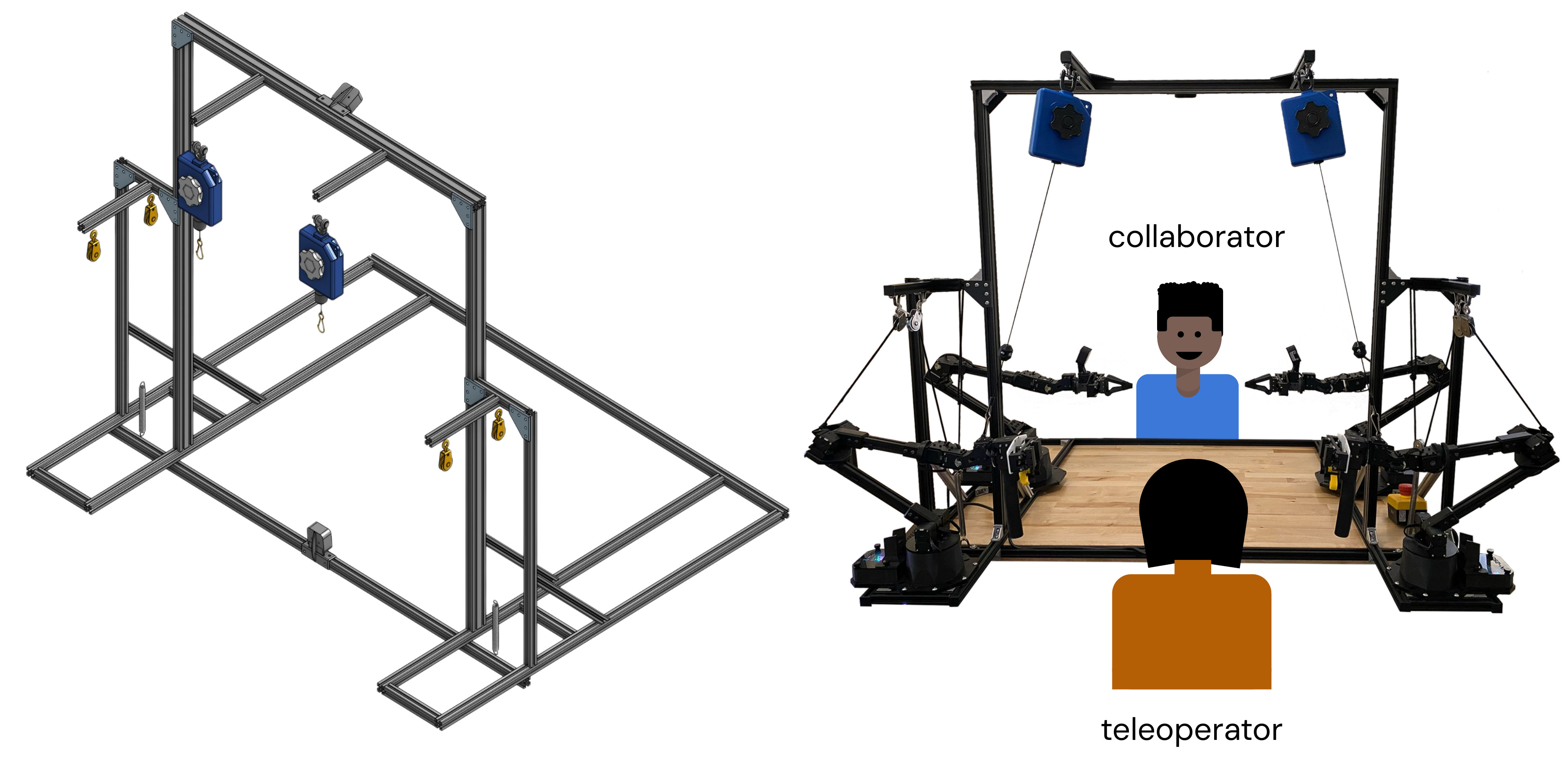}
	\caption{\textit{\textbf{The redesigned frame.}} \textit{\textbf{Left:}} A rendering of the frame, which provides structure for gravity compensation and mount points for the cameras. \textit{\textbf{Right:}} The frame allows space for collection of human-robot collaborative data.}
	\label{fig:topdown_collab}
\end{figure}

We redesign the support frame and build it using 20x20mm aluminum extrusions. The frame provides support for the leader robots, gravity compensation system, and provides mount points for the overhead and worms-eye cameras. Compared to \alohacomma, we simplify the design to remove the vertical frames on the side of the table opposite to the teleoperator. The added space allows for diverse styles of data collection. For example, a human \textit{collaborator} can more easily stand at the opposite side of the workspace and interact with the robot, allowing collection of human-robot interactive data. Additionally, larger props can be placed in front of the table for the robot to interact with.

\subsection{Cameras}

\begin{figure}[t]
	\centering
	\includegraphics[width=\columnwidth]{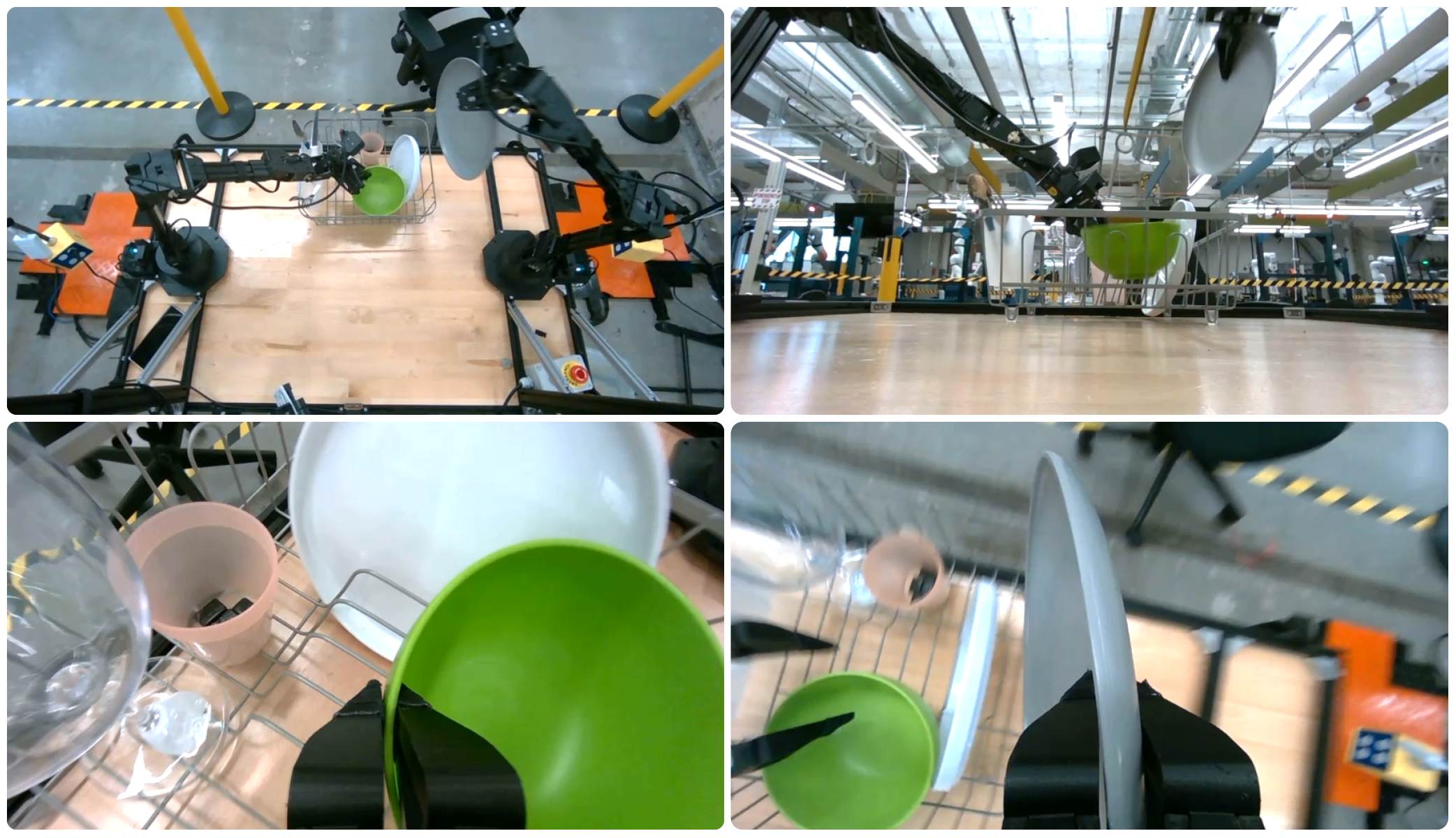}
	\caption{\textit{\textbf{The four camera views recorded during teleoperation on a real workcell.}} From left to right: overhead camera, worms-eye camera, left wrist camera, right wrist camera. All four cameras record 848 x 480 rgb images.}
	\label{fig:camera_views}
\end{figure}

We upgrade the cameras used in the \aloha system to 4 RealSense D405 cameras. These cameras enable high resolution RGB and depth in a small form factor, as well as provide global shutter. We note that although depth and global shutter were not necessary for the results demonstrated on the first \aloha system, they might be considered ``nice to haves'' for enabling different experimentation and pushing performance.  We design new camera mounts for both the wrist cameras, as well as the overhead and worms-eye views (See \figureref{fig:camera_views}). The lower profile of the cameras on the wrists reduces the number of collision states and improves teleoperation for certain fine grained manipulation tasks, especially those that require close contact between the arms or navigating through tight spaces.

\section{Teleoperation}

We run the teleoperation software stack using ROS2 \citep{doi:10.1126/scirobotics.abm6074}. Upon startup, both leader and follower arms initialize to the home position. Operators can start a data collection \textit{session} by closing the gripper using the finger attachments on either of the leader robots. Operators can save or discard a session using foot pedals positioned underneath the workcell.

During a teleoperation session, we log sensor streams from the robot including images, leader and follower joint positions, and other auxiliary data provided by the ROS2 system. We take several measures at collection time to ensure that downstream  pipelines receive complete, high quality data, since this is crucial for robotic learning pipelines. Session statistics such as sensor availability and latency are visible to the operator during collection to ensure data is reliably logged. Sessions are automatically shut down for missing data to ensure downstream learning pipelines always receive complete data. When sessions are logged, we record the operator username, time, and the robot identifier along with the raw sensors streams from the robot. Including the additional data allows filtering data downstream if issues are found for certain robots during a period of time. We record leader and follower joint data at 50Hz, which as the original \aloha system showed, outperforms lower frequency teleoperation.

\pagebreak

\noindent \textbf{Ergonomics.} Much of the motivation for the improvements highlighted in this report is to ensure comfort for users during teleoperation. We reiterate several of the features, and add several new considerations to maximize ergonomic benefit:

\begin{itemize}
  \item \textbf{Low friction grippers} reduce strain on fingers and wrists during teleoperation.
  \item \textbf{Passive gravity compensation} counteracts the weight of the leader arms, to reduce wear on shoulders and arms during long teleoperation sessions.
  \item \textbf{Swappable finger attachments} of different sizes on the leaders allow customization for users with different hand sizes.
  \item \textbf{Adjustable height chairs} allow users to adjust to their optimal height during teleoperation.
  \item \textbf{Suggested rest intervals} for users to take breaks at least 2 minutes long, and avoid long continuous sessions. We observe that taking breaks at least every 10 minutes minimizes wear from repetitive motion. Users are also instructed to take breaks upon feeling any signs of fatigue, and mix up tasks to teleoperate between breaks to avoid continuous repeated motions for a single task.
\end{itemize}

\section{Simulation}

\begin{figure}[h!]
	\centering
	\includegraphics[width=\columnwidth]{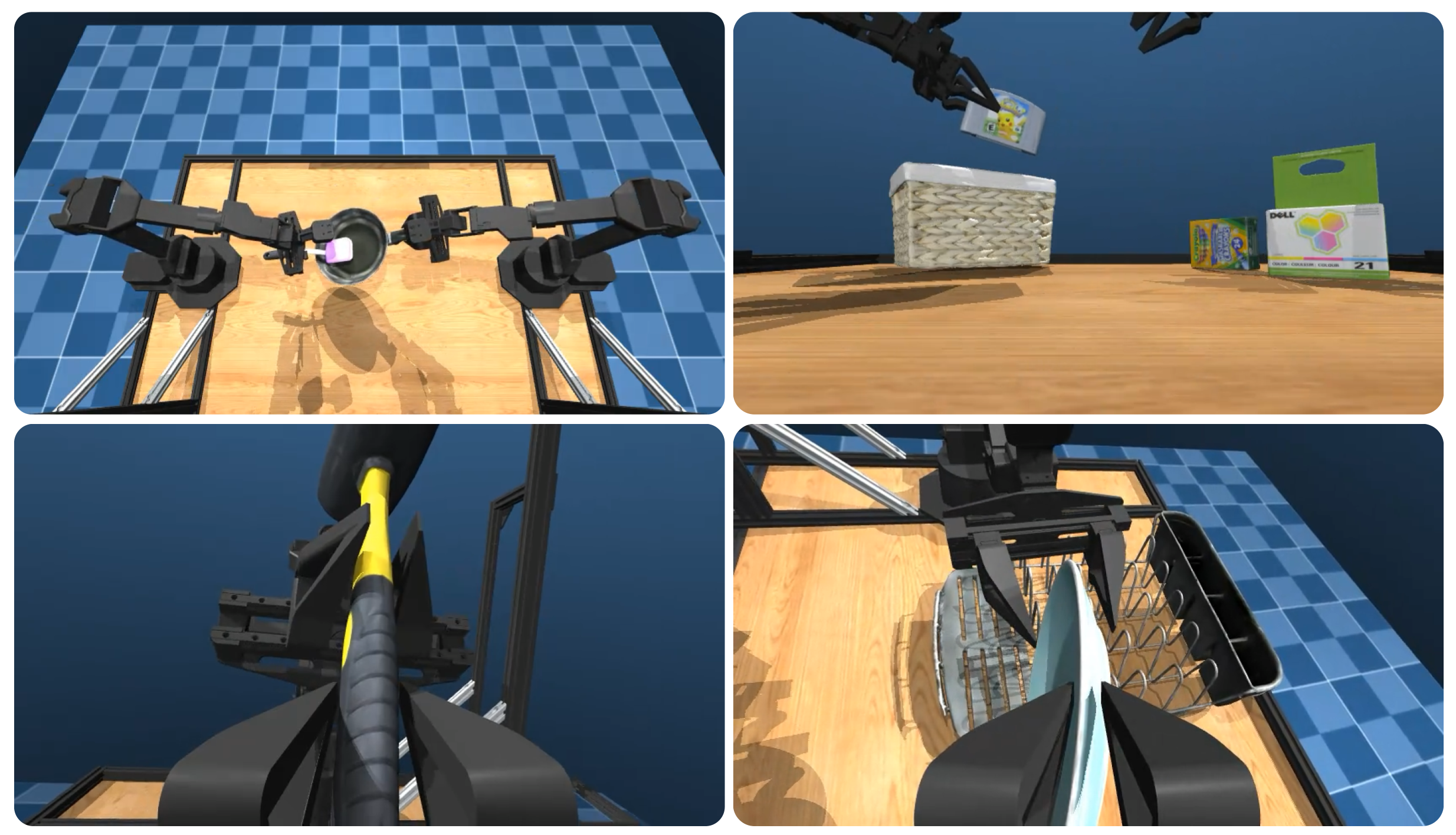}
	\caption{\textit{\textbf{Teleoperating tasks in simulation.}} We show the 4 camera views recorded during simulation collection runs using objects from the Google Scanned Objects Dataset \citep{downs2022scannedobjects}.}
	\label{fig:sim4cams}
\end{figure}

We release a MuJoCo Menagerie \citep{todorov2012mujoco} \citep{menagerie2022github} model of the \name workcell, useful for teleoperation and learning in simulation. The new model is \textbf{more physically accurate} and has \textbf{higher visual fidelity} than previously released \aloha models. We perform system identification using logged trajectories from a real \name setup to set physics parameters in the MuJoCo model. In particular, we collect 11 trajectories in real using the leader arms and minimize the residuals between real and simulated trajectories using a nonlinear least squares solver. Real trajectories consist of sinusoidal motions targeting the control limits of motors in the follower arm. The optimization tunes the proportional gain, damping, armature, joint friction, and torque limits of all position controlled actuators. The gripper is modeled using a position controlled linear actuator with an equality constraint between the gripper fingers. For higher visual fidelity, we match camera intrinsics as closely as possible to the real setup (See \figureref{fig:sim4cams} for the simulated camera views), and we import assets for the table, table extrusions, and follower grippers.

\begin{figure}[h!]
	\centering
	\includegraphics[width=\columnwidth]{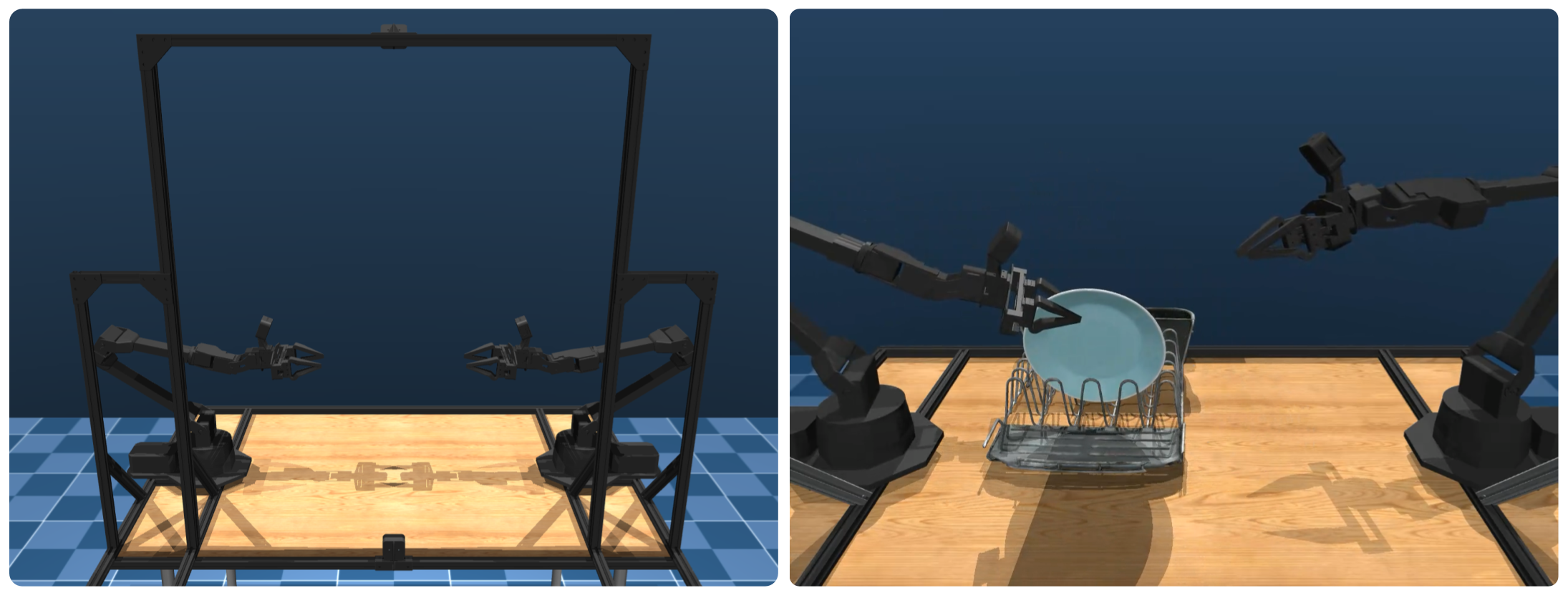}
	\caption{\textit{\textbf{Rendering of the MuJoCo model.}} The model contains ViperX robots with wrist cameras, mounted on a replica of the aluminum extrusion frame. We precisely model all cameras and robot positions of the \name specification, and perform system identification to ensure similar behavior to real.}
	\label{fig:simworkcell}
\end{figure}

The realistic model allows fast, intuitive, and scalable simulation data collection using an \name WidowX leader setup. We hope that an open source, high quality model with system identification can enable sharing of teleoperated simulation data across institutions and accelerate research in policy learning in simulation.

\section{Conclusion}

We present a low-cost system for bimanual teleoperation, enhancing performance, user-friendliness, and robustness compared to the previous \aloha system. We make concrete improvements to hardware such as grippers, gravity compensation, frame, and cameras, while also providing a high quality simulation model. We hope \name will enable large scale data collection for fine-grained bimanual manipulation to advance research in robot learning.

% Bibliography components
\bibliographystyle{abbrvnat}
\nobibliography*
\bibliography{template_refs}

\begin{thebibliography}{8}
\providecommand{\natexlab}[1]{#1}
\providecommand{\url}[1]{\texttt{#1}}
\expandafter\ifx\csname urlstyle\endcsname\relax
  \providecommand{\doi}[1]{doi: #1}\else
  \providecommand{\doi}{doi: \begingroup \urlstyle{rm}\Url}\fi

\bibitem[Downs et~al.(2022)Downs, Francis, Koenig, Kinman, Hickman, Reymann,
  McHugh, and Vanhoucke]{downs2022scannedobjects}
L.~Downs, A.~Francis, N.~Koenig, B.~Kinman, R.~Hickman, K.~Reymann, T.~B.
  McHugh, and V.~Vanhoucke.
\newblock Google scanned objects: A high-quality dataset of 3d scanned
  household items, 2022.
\newblock URL \url{https://arxiv.org/abs/2204.11918}.

\bibitem[Keselman et~al.(2017)Keselman, Iselin~Woodfill, Grunnet-Jepsen, and
  Bhowmik]{keselman2017intel}
L.~Keselman, J.~Iselin~Woodfill, A.~Grunnet-Jepsen, and A.~Bhowmik.
\newblock Intel realsense stereoscopic depth cameras.
\newblock In \emph{Proceedings of the IEEE conference on computer vision and
  pattern recognition workshops}, pages 1--10, 2017.

\bibitem[Macenski et~al.(2022)Macenski, Foote, Gerkey, Lalancette, and
  Woodall]{doi:10.1126/scirobotics.abm6074}
S.~Macenski, T.~Foote, B.~Gerkey, C.~Lalancette, and W.~Woodall.
\newblock Robot operating system 2: Design, architecture, and uses in the wild.
\newblock \emph{Science Robotics}, 7\penalty0 (66):\penalty0 eabm6074, 2022.
\newblock \doi{10.1126/scirobotics.abm6074}.
\newblock URL
  \url{https://www.science.org/doi/abs/10.1126/scirobotics.abm6074}.

\bibitem[Todorov et~al.(2012)Todorov, Erez, and Tassa]{todorov2012mujoco}
E.~Todorov, T.~Erez, and Y.~Tassa.
\newblock Mujoco: A physics engine for model-based control.
\newblock In \emph{2012 IEEE/RSJ International Conference on Intelligent Robots
  and Systems}, pages 5026--5033. IEEE, 2012.
\newblock \doi{10.1109/IROS.2012.6386109}.

\bibitem[{Trossen Robotics}({\natexlab{a}})]{ViperX}
{Trossen Robotics}.
\newblock Viperx 300 robot arm 6dof, {\natexlab{a}}.
\newblock URL
  \url{https://www.trossenrobotics.com/viperx-300-robot-arm-6dof.aspx}.
\newblock Accessed: 2024-01-24.

\bibitem[{Trossen Robotics}({\natexlab{b}})]{WidowX}
{Trossen Robotics}.
\newblock Widowx 250 robot arm 6dof, {\natexlab{b}}.
\newblock URL
  \url{https://www.trossenrobotics.com/widowx-250-robot-arm-6dof.aspx}.
\newblock Accessed: 2024-01-24.

\bibitem[Zakka et~al.(2022)Zakka, Tassa, and {MuJoCo Menagerie
  Contributors}]{menagerie2022github}
K.~Zakka, Y.~Tassa, and {MuJoCo Menagerie Contributors}.
\newblock {MuJoCo Menagerie: A collection of high-quality simulation models for
  MuJoCo}, 2022.
\newblock URL \url{http://github.com/google-deepmind/mujoco_menagerie}.

\bibitem[Zhao et~al.(2023)Zhao, Kumar, Levine, and Finn]{zhao2023learning}
T.~Z. Zhao, V.~Kumar, S.~Levine, and C.~Finn.
\newblock Learning fine-grained bimanual manipulation with low-cost hardware,
  2023.

\end{thebibliography}

\clearpage

\section{Author Contributions}
\label{sec:contributions}

Please cite this work as “ALOHA 2 Team (2024)”.

\begin{itemize}
  \item \textbf{Core Team (core contributors; leading the project effort; leading design, implementation, and/or research on the platform)}: Travis Armstrong, Chelsea Finn, Pete Florence, Spencer Goodrich, Thinh Nguyen, Jonathan Tompson, Ayzaan Wahid, and Tony Zhao.
  \item \textbf{Hardware (working on hardware design and manufacturing; assembly of the system)}: Jorge Aldaco, Kenneth Draper, Pete Florence, Spencer Goodrich, Torr Hage, Thinh Nguyen, Jonathan Tompson, Ayzaan Wahid, and Tony Zhao.
  \item \textbf{Software (software systems to run teleoperation and models; DevOps)}: Jeff Bingham, Sanky Chan, Debidatta Dwibedi, Pete Florence, Spencer Goodrich, Wayne Gramlich, Alexander Herzog, Ian Storz, Jonathan Tompson, Sichun Xu, Ayzaan Wahid, Ted Wahrburg, Sergey Yaroshenko, and Tony Zhao.
  \item \textbf{Data (software and infrastructure to collect and process data)}: Robert Baruch, Pete Florence, Jonathan Hoech, Ian Storz, Ayzaan Wahid, and Sergey Yaroshenko.
  \item \textbf{Simulation (working on creating and improving the simulation model)}: Baruch Tabanpour, Ayzaan Wahid, and Kevin Zakka.
  \item \textbf{HRI / User studies (working on HRI, ergonomics, and conducting user studies)}: Travis Armstrong, Spencer Goodrich, Leila Takayama, and Jonathan Tompson.
\end{itemize}

\subsection{Acknowledgements}

We thank Tom Erez, Matthew Mounteer, Francesco Romano, Stefano Saliceti, and Yuval Tassa for help with the simulation model. We thank Tomas Jackson for photography of the \name fleet. We thank Corey Lynch for help with software setup and ROS2 integration. We thank Chikezie Ejiasi for creating character illustrations used in figures.
We would also like to thank members of the wider Google DeepMind Robotics team for their support.

\end{document}